\documentclass[sigplan,nonacm]{acmart}

\graphicspath{ {./fig/} }
\usepackage[greek, english]{babel} 
\usepackage{units} 
\usepackage{subcaption} 
\usepackage{varioref} 
\usepackage{listings} 
\usepackage{cancel} 
\usepackage{enumitem} 

\newcommand{\real}{\mathbb{R}} 
\DeclareMathOperator{\asin}{arcsin} 
\DeclareMathOperator{\atan}{arctan} 
\allowdisplaybreaks[1]

\newcommand{\noun}[1]{\textsc{#1}} 

\theoremstyle{remark}
	\newtheorem*{note}{Note}

	\newtheorem*{problem*}{Problem}
	\newtheorem*{prop*}{Proposition}
	
	\newtheorem*{example*}{Example}
\theoremstyle{definition}

\theoremstyle{plain}

\lstdefinelanguage{GLSL}
{
	morekeywords={
		attribute,const,uniform,varying,
		break,continue,do,
		for,while,
		switch,case,default,
		if,else,
		in,out,inout,
		float,int,void,bool,true,false,
		discard,return,
		mat2,mat3,mat4,
		mat2x2,mat2x3,mat2x4,
		mat3x2,mat3x3,mat3x4,
		mat4x2,mat4x3,mat4x4,
		vec2,vec3,vec4,
		ivec2,ivec3,ivec4,
		bvec2,bvec3,bvec4,
		uint,uvec2,uvec3,uvec4,
		lowp,mediump,highp,precision,invariant,
		sampler1D,sampler2D,sampler3D,samplerCube,
		struct,
		define,undef,
		ifdef,ifndef,elif,endif,
	},
	sensitive=false, 
	morecomment=[l]{//},
	morecomment=[s]{/*}{*/},
	morestring=[b]"
}
\received{November 2024}
\received{December 2024} 

\begin{document}
\title{Lens Distortion Encoding System\\ Version 1.0}

\author{Jakub Maksymilian Fober}
\orcid{0000-0003-0414-4223}
\email{jakub.m.fober@protonmail.com} 
\email{talk@maxfober.space}

\renewcommand\shortauthors{Fober, J.M.}

\begin{abstract}
Lens Distortion Encoding System (LDES) allows for a distortion--accurate workflow, with a seamless interchange of high quality motion picture images regardless of the lens source. This system is similar in a concept to the Academy Color Encoding System (ACES)\cite{Oscars2015ACES}, but for distortion.
Presented solution is fully compatible with existing software/plug-in tools for STMapping found in popular production software like Adobe After Effects or DaVinci Resolve.
LDES utilizes common distortion space and produces single high-quality, animatable STMap used for direct transformation of one view to another, neglecting the need of lens-swapping for each shoot.
The LDES profile of a lens consist of two elements;
\textit{View Map} texture, and \textit{Footage Map} texture, each labeled with the FOV value.
Direct distortion mapping is produced by sampling of the \textit{Footage Map} through the \textit{View Map}. The result; animatable mapping texture, is then used to sample the footage to a desired distortion.
While the \textit{Footage Map} is specific to a footage, \textit{View Maps} can be freely combined/transitioned and animated, allowing for effects like smooth shift from anamorphic to spherical distortion, previously impossible to achieve in practice.
Presented LDES Version 1.0 uses common 32-bit STMap format for encoding, supported by most compositing software, directly or via plug-ins. The difference between standard STMap workflow and LDES is that it encodes absolute pixel position in the spherical image model.
The main benefit of this approach is the ability to achieve a similar look of a highly expensive lens using some less expensive equipment in terms of distortion. It also provides greater artistic control and never seen before manipulation of footage.
\end{abstract}

\ccsdesc[500]{Computing methodologies~Computer vision}
\ccsdesc[500]{Computing methodologies~Image processing}
\ccsdesc[500]{Computing methodologies~Image manipulation}
\ccsdesc[300]{Computing methodologies~Perception}
\ccsdesc[300]{Applied computing~Media arts}
\ccsdesc[100]{Computing methodologies~Ray tracing}


\keywords{
	Camera calibration,
	Lens distortion correction,
	Image processing,
	Anamorphic lenses,
	Aximorphic distortion,
	Fisheye lenses,
	STMap workflow,
	Visual sphere mapping
}

\maketitle


\begin{figure}[bh]
	\footnotesize
	\copyright\ 2024 Jakub Maksymilian Fober\smallskip\\
	\href{https://creativecommons.org/licenses/by-nc-nd/3.0/}{\includegraphics{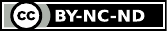}}\\
	This work is licensed under Creative Commons BY-NC-ND 3.0 license.
	\url{https://creativecommons.org/licenses/by-nc-nd/3.0/legalcode}\\
	For inquiries about custom licensing or collaboration opportunities, please contact the author.
\end{figure}


\section{Introduction}\label{sec:introduction}

Typical STMap workflow usually involves a mapping between rectilinear projection and specific lens distortion. It is used for image--to--motion camera solving and compositing of Computer Generated Images (CGI) with real lens footage. The typical process of STMap workflow is limited to a specific camera setup and does not allow for a third-party lens profile end-result. The tools are widely available in compositing and 3D imaging software, both for big-budget production as well as low--budget free software. STMapping is de-facto standard for exchangeable lens distortion management in the video--and--film production industry.\cite{Houdini2024STMaps}

\subsection[New System Capability]{New System Allows for Effects Never Seen Before}\label{subsec:new possibilities}

Beside achieving all the goals of STMap workflow, the LDES solution gives in hands of artists an unlimited digital lens, not constrained by mechanical limitations of optical assembly.\cite{Sasaki2017_5PilarsOfAnamorphic} It can smoothly transition between different lens types, parameters, and utilize synthetic projections.\cite{Williams2017Minecraft} It is not constrained by the FOV limit of rectilinear projection, but can work with 0--360\textdegree\ footage.
This is possible thanks to transition from rectilinear--pyramidal image base model to spherical model.\cite{Alberti1970OnPaining} Which encapsulates all possible perspective projections.

The notion of sphere as a projective surface, a base model for the concept of image was already theorized in ancient times\cite{McArdle2013VisualSphereEuclid}. Even doe rectilinear projection is an old standard, at its conception, one of the most famous artists, \noun{Leonardo DaVinci} was known to criticize it for unnatural presentation of proportions.\cite{Dixon1987Paradox,Argan1946BrunelleschiPerspective,DaVinci1632TeatriseOnPainting} Today we know that pictures are projections of visual sphere and exhibit same issues found in a field of cartography.\cite{Penaranda2015SphereInterpolation,Sharpless2010Panini,German2007PanoramaOnFlatSurface,Fleck1994FisheyeEquations} The knowledge of the principles found in map projection is useful in creating artistically conscious decisions about desired lens distortion. It can lead an artist to direct viewers perception for the photographed space and its aspects, such that end result is pleasing to the eye and feels natural.
Special projection parameters can be tailored to give accurate notion of spatial dimension, speed, distance or proportions to the viewer.
This is achieved with LDES and synthetic projections workflow.
\pagebreak


\subsection[Mathematical Convention]{Document Mathematical Convention}

This document adheres to the following naming conventions:
\begin{itemize}[label=$\star$]
    \item A left-handed coordinate system is used.
    \item Vectors are presented in column format.
    \item Matrices follow a row-major order and are denoted as ``$M_{\text{row}\,\text{col}}$''.
    \item Matrix multiplication is represented as ``$[\text{column}]_a \cdot [\text{row}]_b = M_{a\,b}$''.
    \item A single bar enclosure ``$|u|$'' applied to a scalar denotes its \textit{absolute value}.
    \item A single bar enclosure ``$|\vec v|$'' applied to a vector denotes its \textit{length} or \textit{magnitude}.
    \item Vectors, whether signed or unsigned, are computed component-wise to produce another vector.
    \item The centered dot ``$\cdot$'' indicates the \textit{dot product} of two vectors.
    \item Square brackets with a comma ``$[f, c]$'' denote an \textit{interval}.
    \item Square brackets without a comma ``$[x\ y]$'' denote a \textit{vector} or a \textit{matrix}.
    \item The exponent ``$^{-1}$'' signifies the \textit{reciprocal} of the given value.
    \item The \emph{QED} symbol ``$\square$'' marks the final result or output.
\end{itemize}
These conventions streamline the translation of mathematical formulations into code.

\section{Lens Distortion Correction Basics}\label{sec:lens basics}

In this section, we will discuss the types of lens distortions and provide a brief overview of their correction methods. For a more detailed explanation of the mathematical foundations, refer to section No. \vref{sec:math}.

\subsection{Lens Distortions}\label{subsec:distortion types}

Geometric lens distortions can be categorized into two primary types: projection-based distortions and aberrations. Aberrations include barrel distortion, pincushion distortion, and a combination of the two, known as mustache (or radial) distortion. Other types include decentering distortion, which arises from misalignment of optical elements, and thin prism distortion, caused by the tilt of an optical element relative to the imaging sensor.\cite{Wang2008BrownConradyNewModelLensDistortion}

In addition to geometric distortions, there are non-geometric aberrations such as vignetting, which darkens the image toward its periphery; chromatic aberration, which splits incoming light into a spectrum; bokeh, which affects the appearance of out-of-focus areas; and lens flares.

This document primarily focuses on geometric distortions and geometric aberrations, providing a limited overview of vignetting as caused by projection geometry. It is worth noting that vignetting can also result from the occlusion of light by the lens casing, an effect most pronounced when using wide aperture settings and is specific to a given lens setup.

Projection-based distortions result from the projection of the visual sphere onto a flat surface and can be categorized using cartographic terms. In motion picture applications, azimuthal projections are particularly useful as they exhibit only radial distortion, preserving the straightness of lines directed toward the vanishing point on the optical axis.
Equirectangular projection can also be beneficial during the production stage, as it encapsulates the entire visual sphere.

Anamorphic or aximorphic distortions represent a special case where the distortion profile differs between the vertical and horizontal axes of the image. Anamorphic lenses are particularly popular in film, commercials, and music video production due to their unique aesthetics when the image is in motion.\cite{Sasaki2017Anamorphic,Neil2004CineLens} In contrast, they are rarely used in photography, which is predominantly characterized by wide-angle rectilinear lenses.\cite{Kingslake1989RapidRectilinearLens}

These distortions can arise from projection-based effects or may exhibit axis-dependent aberrations. Spherical fisheye lenses, which belong to the family of cartographic projections, are primarily used in sports and action shots where extensive all-axis camera motion is common.\cite{Thoby2012FisheyeLensComparison}

\subsection{Correction Methods}\label{subsec:correction models}

The typical STMap workflow focuses on aberration correction and mapping to rectilinear projection. The most commonly used method in this case is the \textit{Brown-Conrady model}, which addresses various radial aberrations and optical misalignments.\cite{Wang2008BrownConradyNewModelLensDistortion} This model operates entirely in image space and is independent of the camera's field of view (FOV).

However, the Brown-Conrady method has a limitation when correcting for severe barrel distortion, such as those found in fisheye lenses. In such cases, cartographic mapping provides a more effective solution.\cite{Krause2019PTGuiFisheye_k-factor}

Since the LDES is based on visual sphere mapping and is FOV-dependent, cartographic mapping is the preferred approach. When combined with the Brown-Conrady division model, this method enables comprehensive mapping of all lenses.
For anamorphic and aximorphic distortions, a custom extended version of this model is employed, allowing individual control over the distortion for each axis of the image.\cite{Fober2021Aximorphic,Yuan2009AnamorphicLens}

\subsection{Calibration Methods}\label{subsec:calibration}

The LDES provides two calibration methods for obtaining the \textit{view map} and \textit{footage map} of a real lens.

The first method involves fitting the projection model to the lens using checkerboard calibration.

The second method directly maps points in the frame to incidence angles using a specialized calibration rig. This rig includes a visible target at infinity (e.g., a collimated dot--sight) and a 2-axis motorized gimbal. The software tracks the position of the target in the frame and correlates it with the corresponding pan and tilt positions of the gimbal, thereby determining the incidence direction for a given frame position. Additionally, the sparse data obtained from this method can be used to fit the projection model, offering a hybrid approach to calibration.

\section{Lens Distortion Encoding System}\label{sec:system}

The Lens Distortion Encoding System (LDES) incorporates a specialized type of STMaps, that seamlessly combine various lens profiles within a unified encoding scheme. This approach enables the creation of a lens distortion library, providing flexibility of choice in achieving the desired look and feel of the processed image, regardless of its source.

\subsection{System Overview}\label{subsec:system overview}

The system relies on two primary texture map types, both associated with a single lens profile. This separation enables direct mapping from one lens to another through a synthesized texture map. During this process, multiple distortion profiles can be combined and animated, allowing for smooth transition between lenses—a technique that was previously nearly impossible to achieve.

Additionally, the LDES supports the application of exotic distortion profiles and the compositing of sources that traditionally where incompatible. While maintaining high fidelity of the final image, the system provides unprecedented creative flexibility and precision in lens distortion handling.

\subsection{Core Principles}\label{subsec:core principles}

The system utilizes two types of STMapping files, each corresponding to a single lens profile. The second type can be derived from the first.

\smallskip
\textit{View maps} match the dimensions of the footage, preserving the original resolution and aspect ratio. The blue channel of the texture can optionally encode vignetting in linear space. This encoding allows for correction of source footage using a division blending mode or the addition of vignetting to the final image using a multiplication blending mode. \textit{View maps} do not encode masking in the alpha channel. The coordinates of \textit{view maps} map to the equidistant projection space of a specified FOV value.

\smallskip
\textit{Footage maps} are always square textures and do not encode vignetting in the blue channel. Instead, the alpha channel of the \textit{footage map} visualizes the boundaries of the footage in equidistant space. The coordinates of \textit{footage maps} map directly to the footage.

\smallskip
\textit{View maps} from different sources can be combined using opacity blending, and then applied to a footage of a third source type.

\subsection{Implementation Details}\label{subsec:implementation}

Texture maps are encoded in a 32-bit floating-point format, such as \emph{TIFF} or \emph{OpenEXR}. These file-types ensure high precision and compatibility across different software. Each texture map filename should adhere to the specified naming convention:

\begin{figure}[h]
\textit{FOV values are written as whole numbers, representing the horizontal axis in degrees.}
	\begin{equation*}
		\underbrace{\texttt{ViewMap}}_\text{type}
		\texttt{\_}
		\underbrace{\texttt{Cinerama1950}}_\text{description}
		\texttt{\_}
		\underbrace{\texttt{FOV146}}_\text{FOV}
		\texttt{.tif}
	\end{equation*}
	\begin{equation*}
		\underbrace{\texttt{FootageMap}}_\text{type}
		\texttt{\_}
		\underbrace{\texttt{TTArtisan7.5mm\_Fisheye}}_\text{description}
		\texttt{\_}
		\underbrace{\texttt{FOV180}}_\text{FOV}
		\texttt{.tif}
	\end{equation*}
\textit{An additional prefix ``n'' is added for normalized FOV values of view map filenames. For more details, see equation No. \vref{eq:FOV normalization}.}
	\begin{equation*}
		\underbrace{\texttt{ViewMap}}_\text{type}
		\texttt{\_}
		\underbrace{\texttt{Cooke14mm\_WideAngle}}_\text{description}
		\texttt{\_}
		\underbrace{\texttt{nFOV92}}_\text{alt. FOV}
		\texttt{.tif}
	\end{equation*}
	\caption[Naming convention]{Naming convention for the LDES files.}
	\label{fig:naming convention}
\end{figure}

The inclusion of the FOV value in the filename is mandatory, for the mapping process to function. The FOV is always specified as a whole number in degrees. If the exact lens value is a decimal numeral, the texture coordinates must be scaled appropriately to match the noted FOV value. For further details, see equation No. \vref{eq:FOV normalization}. As a good practice, the FOV noted value should be rounded up in the creation process.

\section{Mathematical Foundations}\label{sec:math}

Both \textit{view maps} and \textit{footage maps} encode STMap coordinates. The \textit{footage map} represents footage $ST$ coordinates in equidistant space $\theta$. The boundary of the \textit{footage map} corresponds to an FOV angle of $\nicefrac{\Omega}{2}$.

\begin{equation}
\label{eq:footage map}
	\begin{bmatrix}
		\vec{\text{f\_map}}_s \\
		\vec{\text{f\_map}}_t
	\end{bmatrix}
	\mapsto
	\begin{bmatrix}
		\vec{\text{tex}}_s \\
		\vec{\text{tex}}_t
	\end{bmatrix}
\end{equation}

The \textit{view map} $ST$ coordinates encode the \textit{footage map} coordinates within the image space of the result. These coordinates are normalized to a specified FOV angle $\Omega$.

\begin{subequations}
\label{eq:view map}
\begin{align}
	\begin{bmatrix}
		\vec{\text{v\_map}}_s \\
		\vec{\text{v\_map}}_t
	\end{bmatrix}
	\mapsto
	\begin{bmatrix}
		\vec{\text{f\_map}}_s \\
		\vec{\text{f\_map}}_t
	\end{bmatrix}
\\
	\Omega_{\text{v\_map}}
	\left|
	\begin{bmatrix}
		\vec{\text{v\_map}}_s \\
		\vec{\text{v\_map}}_t
	\end{bmatrix}
	-\frac{1}{2}
	\right|
	=
	\theta
\end{align}
\end{subequations}

If there is a mismatch between the \textit{footage map} and \textit{view map} FOV angle $\Omega$, a simple tile scaling is applied when sampling the \textit{footage map} over the \textit{view map} coordinates.

\begin{equation}
\label{eq:tile scaling}
	\text{tile\_scale} = \frac{\Omega_{\text{view\_map}}}{\Omega_{\text{footage\_map}}}
\end{equation}

Both maps are not intended to directly sample the footage, as this would result in quality loss. Instead, the \textit{footage map} is sampled through the \textit{view map} to produce the final distortion mapping STMap, which is then used to sample the footage.

\begin{equation}
\label{eq:sampling}
	\begin{bmatrix}
		\vec{\text{v\_map}}_s \\
		\vec{\text{v\_map}}_t
	\end{bmatrix}
	\overset{\textit{samples}}{\rightarrow}
	\begin{bmatrix}
		\vec{\text{f\_map}}_s \\
		\vec{\text{f\_map}}_t
	\end{bmatrix}
	\quad\mapsto
	\begin{bmatrix}
		\vec{\text{tex}}_s \\
		\vec{\text{tex}}_t
	\end{bmatrix}
\end{equation}

\paragraph{Animated View Maps}\label{par:animation}

View maps can be combined to create animated \textit{view maps} that transition between different distortion parameters. This animation is achieved by interpolating between the two maps, for example, by adjusting the opacity of one map over the other.

To ensure a seamless transition, it is recommended to first match the FOV value $\Omega$ of each processed \textit{view map}, avoiding necessity for tile-scaling during animation. Normalization of the encoded FOV angle is performed by scaling the \textit{view map} coordinates to a common FOV value, which then is used to control the tile scaling.

\begin{equation}
\label{eq:FOV normalization}
	\begin{bmatrix}
		\vec{\text{v\_map}'}_s \\
		\vec{\text{v\_map}'}_t
	\end{bmatrix}
	=
	\frac{\Omega_{\text{v\_map}}}{\Omega_{\text{common}}}
	\left(
	\begin{bmatrix}
		\vec{\text{v\_map}}_s \\
		\vec{\text{v\_map}}_t
	\end{bmatrix}
	-\frac{1}{2}
	\right)
	+\frac{1}{2}
\end{equation}

\paragraph{Ray-Tracing with Lens Distortion}\label{par:ray-tracing}

View maps can be utilized to directly drive the ray-tracing engine in 3D imaging software, such as \noun{Blender}, by transforming the encoded coordinates into incidence $\hat{I}$ vectors.\cite{BlenderManual2024RayPortalBSDF}

\begin{subequations}
\label{eq:ray map}
\begin{align}
	\theta &=
	\Omega_{\text{v\_map}}
	\left|
	\begin{bmatrix}
		\vec{\text{v\_map}}_s \\
		\vec{\text{v\_map}}_t
	\end{bmatrix}
	-\frac{1}{2}
	\right|
\\
	\begin{bmatrix}
		\hat I_x \\
		\hat I_y \\
		\hat I_z
	\end{bmatrix}
	&=
	\begin{bmatrix}
		\sin(\theta)
		\left\|
		\begin{bmatrix}
			\vec{\text{v\_map}}_s \\
			\vec{\text{v\_map}}_t
		\end{bmatrix}
		-\frac{1}{2}
		\right\|
		\\
		\cos(\theta)
	\end{bmatrix}
\end{align}
\end{subequations}

\paragraph{Rotating View Map Camera}

View map coordinates can be rotated to simulate camera pan, tilt, and roll motion effects. This is achieved by mapping the \emph{view map} coordinates to incidence vectors, which can then be rotated.
\begin{equation}
	\vec{\text{v\_map}}\mapsto
	\theta\mapsto
	\hat{I}
	\overset{\textit{rotation}}{\rightarrow}
	\hat{I}'\mapsto
	\theta'\mapsto
	\vec{\text{v\_map}}'
\end{equation}

\subsection{Distortion Models}\label{subsec:models}

The LDES employs a cartographic fisheye projection model as the foundation for synthetic view maps.\cite{Krause2019PTGuiFisheye_k-factor} These maps can be further refined through extensions to align with the distortion profile of real-life lens.

\paragraph{Synthetic Projection Base Model}\label{par:synthetic model}

This model allows for a seamless transition from rectilinear projection to a fisheye-orthographic view. Each value of $k$ below 1 corresponds to a different type of fisheye lens. The table No. \vref{tab:k values} outlines the major values and their corresponding lens type.

\begin{table}[h]
	\centering
	\begin{tabular}{rrl}
		\toprule
		\multicolumn{2}{l}{Value of $k$} & Fisheye Type \\ \midrule
		$k\;=$ & $1$                & Rectilinear \textit{(Gnomonic)} \\
		$k\;=$ & $\nicefrac{1}{2}$  & Stereographic \\
		$k\;=$ & $0$                & Equidistant \\
		$k\;=$ & $-\nicefrac{1}{2}$ & Equisolid \\
		$k\;=$ & $-1$               & Orthographic \textit{(azimuthal)} \\
		\bottomrule
	\end{tabular}\\
	\footnotesize\emph{Source:} PTGui 11 fisheye factor \cite{Krause2019PTGuiFisheye_k-factor}.
	\smallskip
	\caption{Primary $k$ values and corresponding fisheye projection type.}
	\label{tab:k values}
\end{table}

\begin{subequations}
\label{eq:spherical theta}
\begin{align}
	f &=
		\begin{cases}
			\frac{\tan\big(\frac{\Omega}{2}k\big)}{k}, & \text{if } k>0 \\
			\frac{\Omega}{2}, & \text{if } k=0 \\
			\frac{\sin\big(\frac{\Omega}{2}k\big)}{k}, & \text{if } k<0 \\
		\end{cases}
\\
	r &= |\vec v| = \sqrt{\vec v^2_x+\vec v^2_y} =
		\begin{cases}
			\frac{\tan(\theta k)f}{k}, & \text{if } k>0 \\
			\theta f, & \text{if } k=0 \\
			\frac{\sin(\theta k)f}{k}, & \text{if } k<0 \\
		\end{cases}
\\
	\theta &=
		\begin{cases}
			\frac{\atan\big( \frac{r}{f}k \big)}{k}, & \text{if } k>0 \\
			\frac{r}{f}, & \text{if } k=0 \\
			\frac{\asin\big( \frac{r}{f}k \big)}{k}, & \text{if } k<0 \\
		\end{cases}
\end{align}
\end{subequations}
where:
\begin{itemize}[label=$\star$]
	\item $f$ is the focal length derived from horizontal FOV angle $\Omega$,
	\item $r$ is the image space radius originating from the center, which is derived from:
	\item $\vec v$ the view coordinate $\real^2$ vector in image space, with correct aspect ratio, originating from the center.
\end{itemize}

\paragraph{Anamorphic Extension}\label{par:anamophic}

The anamorphic extension to the radius $r$ enables vertical scaling of the distortion based on the anamorphic squeeze factor $s$.

\begin{subequations}
\label{eq:anamorphic}
\begin{align}
	r &= \sqrt{\vec v^2_x+\frac{\vec v^2_y}{s}}
\\
	\vec v' &= \left|\frac{\vec v}{r}\right|\vec v
\end{align}
\end{subequations}

\paragraph{Aximorphic Extension}\label{par:aximorphic}

The aximorphic extension to the $\theta$ angle provides independent control over the projection type for the vertical and horizontal axes. It serves as an alternative to anamorphic projection. The angles $\theta_x$ and $\theta_y$ are calculated separately, each with its own $k$ value, while sharing a common focal length $f$.

\begin{subequations}
\label{eq:aximorphic interpolation}
\begin{align}
	\begin{bmatrix}
		\vec w_x \\
		\vec w_y
	\end{bmatrix}
	&=
	\begin{bmatrix}
		\cos^2\varphi \\
		\sin^2\varphi
	\end{bmatrix}
	=
	\begin{bmatrix}
		\frac{1}{2}+\frac{\cos(2\varphi)}{2} \smallskip\\
		\frac{1}{2}-\frac{\cos(2\varphi)}{2}
	\end{bmatrix}
	=
	\begin{bmatrix}
		\frac{\vec v^2_x}{\vec v^2_x+\vec v^2_y} \smallskip\\
		\frac{\vec v^2_y}{\vec v^2_x+\vec v^2_y}
	\end{bmatrix}
\\
	\theta'
	&=
	\vec w\cdot\vec\theta_{xy}
	=
	\vec w_x\vec\theta_x+\vec w_y\vec\theta_y
\end{align}
\end{subequations}
where:
\begin{itemize}[label=$\star$]
	\item $\vec w$ consists of aximorphic interpolation weights,
	\item $\varphi$ is the azimuthal angle of the incidence vector in the visual sphere space,
	\item $\theta$ is the polar angle of the incidence in the visual sphere space.
\end{itemize}

\paragraph{Asymmetrical-Aximorphic Extension}\label{par:assymetrical-aximorphic}

The aximorphic projection can be further extended to provide independent control over the top and bottom portions of the image with respect to the projection type.

\begin{equation}
\label{eq:asymmetry}
	k_y =
		\begin{cases}
			k_\textit{top}, & \text{if }\vec v_y>0 \\
			k_\textit{bottom}, & \text{if }\vec v_y\leq0
		\end{cases}
\end{equation}

\paragraph{Brown-Conrady Extension}\label{par:brown-conrady}

The Brown-Conrady extension offers geometric aberration control over the distortion profile. It can be seamlessly integrated with both anamorphic and aximorphic extensions.

\begin{subequations}
\label{eq:brown-conrady}
\begin{align}
	\begin{bmatrix}
		\vec f_x \\
		\vec f_y
	\end{bmatrix}
	&=
	\underbrace{
		\begin{bmatrix}
			\vec v_x-c_1 \\
			\vec v_y-c_2
		\end{bmatrix}
	}_\text{cardinal offset (A)}
	\\
	r^2 &= \vec f^2_x+\vec f^2_y
	\\
	\begin{split}
		\begin{bmatrix}
			\vec v'_x \\
			\vec v'_y
		\end{bmatrix}
	&=
		\begin{bmatrix}
			\vec f_x \\
			\vec f_y
		\end{bmatrix}
		\underbrace{
			\big(
				1+k_1r^2+k_2r^4+k_3r^6\cdots
			\big)^{-1}
		}_\text{radial}
		+
		\underbrace{
			\begin{bmatrix}
				c_1 \\
				c_2
			\end{bmatrix}
		}_\text{cardinal offset (B)} \square
	\\
	&+
		\underbrace{
			\begin{bmatrix}
				\vec f_x \\
				\vec f_y
			\end{bmatrix}
			\left(
				\begin{bmatrix}
					\vec f_x \\
					\vec f_y
				\end{bmatrix}
				\cdot
				\begin{bmatrix}
					p_1 \\
					p_2
				\end{bmatrix}
			\right)
		}_\text{decentering}
		+
		\underbrace{
			r^2
			\begin{bmatrix}
				q_1 \\
				q_2
			\end{bmatrix}
		}_\text{thin prism} \qed
	\end{split}
\end{align}
\end{subequations}
where:
\begin{itemize}[label=$\star$]
	\item $c_1$ and $c_2$ are the cardinal offset parameters for horizontal and vertical axis,
	\item $k_1$, $k_2$, $k_n$ are the radial distortion coefficients (different from the $k$ projection parameter),
	\item $p_1$ and $p_2$ are the decentering parameters,
	\item $q_1$ and $q_2$ are the thin prism distortion parameters.
\end{itemize}

\paragraph{Aximorphic Brown-Conrady Extension}\label{par:aximorphic-brown-conrady}

The Brown-Conrady extension can be further enhanced to provide independent control over radial distortion for the vertical and horizontal axes. This is particularly useful for matching anamorphic lens distortion.

\begin{equation}
\label{eq:brown-conrady aximorphic}
	\begin{bmatrix}
		\vec f_x \\
		\vec f_y
	\end{bmatrix}
	\underbrace{
		\Bigg(
		\begin{bmatrix}
			\vec w_x \\
			\vec w_y
		\end{bmatrix}
	\cdot
		\begin{bmatrix}
			1+\vec k_{x1}r^2+\vec k_{x2}r^4+\vec k_{x3}r^6\cdots
			\\
			1+\vec k_{y1}r^2+\vec k_{y2}r^4+\vec k_{y3}r^6\cdots
		\end{bmatrix}
		\Bigg)^{-1}
	}_\text{aximorphic radial}
\end{equation}
where:
\begin{itemize}[label=$\star$]
	\item $\vec k_{xn}$ and $\vec k_{yn}$ are same as defined before, but driving horizontal and vertical axis separately.
\end{itemize}

\subsection{Interpolation and Sampling}\label{subsec:interpolation sampling}

Bilinear interpolation can be applied throughout the entire workflow with satisfactory results. However, \textit{view map} sampling of the \textit{footage map} may benefit from bi-cubic Catmull-Rom interpolation, although it is not strictly necessary.

Due to the nature of STMaps, which consist of smooth gradients, high resolution is generally unnecessary. Conversely, the final mapping STMap can benefit from resolution scaling beyond the size of the original footage. This is particularly relevant when using third-party plugins for STMapping in compositing software, as some of these plugins lack support for MIP-mapping or anisotropic filtering. In such cases, additional resolution scaling of the STMap can act as a multi-sampling filter, mitigating potential quality loss in the final distorted image.

\section{Workflow Example}\label{sec:workflow}

In this section, we will explore two types of workflows: the application of LDES maps to a footage and the process of selecting a synthetic distortion profile for a given scene.

\subsection{Compositing Example}\label{subsec:compositing example}

In this thought exercise, we will examine the compositing process for combining Computer Generated Images (CGI) with real camera footage, viewed through an animated synthetic projection lens.

\paragraph{Overview of Assets}\label{par:assets}

We begin with high-resolution footage shot using a wide fisheye lens. Calibration assets include a \textit{view map} with measured vignetting and a derived \textit{footage map}. For the final output, we select two synthetic \textit{view maps} from a library. Additionally, a rectilinear \textit{view map} is generated for solving camera motion from the footage. Below is a summary of the assets:

\begin{enumerate}
\label{lst:assets}
    \item High-resolution live footage captured with a wide-angle fisheye lens.
    \item Calibration \textit{view map} corresponding to the footage, including vignetting.
    \item Calibration \textit{footage map} matching the footage.
    \item Synthetic \textit{view map} resembling an anamorphic lens.
    \item Synthetic \textit{view map} resembling a spherical lens with the same FOV.
    \item Rectilinear \textit{view map} for solving camera motion.
    \item 3D scene for rendering.
\end{enumerate}

The compositing software used is \textit{Adobe After Effects} with the \textit{PixMap} plug-in.\cite{Wunkolo2024PixMap} All assets are imported into the program, ensuring that color management is disabled for the maps. The composition is set to 32-bit floating-point precision to maintain the proper image quality. The project color Working Space is configured as \emph{HDTV (Rec. 709)} with enabled Linearization of Working Space. This setup ensures correct vignette blending.

\paragraph{Tracking Rectified Footage}\label{par:camera tracking example}

Next, we rectify the footage and track the camera motion for use in our 3D software. To begin, we create a composition using the rectilinear \textit{view map} and import the footage along with its corresponding \textit{footage map}. The \textit{PixMap} effect is applied to the rectilinear \textit{view map} layer. We flip the vertical tiling and adjust the tiling scale to account for the FOV mismatch between the \textit{view map} and the \textit{footage map} (see equation No. \vref{eq:tile scaling}).

The sampled texture is then set to the \textit{footage map} layer, followed by applying a second \textit{PixMap} effect, which samples the footage layer. The result is a layer with rectified footage ready for camera tracking.

Before proceeding to render the 3D scene, we can prepare the animation of the lens distortion, which will later be used for ray-tracing the scene. This approach eliminates the need to warp the CGI sequence to match it with the final output.

\paragraph{Animating View Maps}\label{par:animation example}

To animate the two synthetic \textit{view maps}, we create a new composition matching the \textit{view map} size and import both maps. The distortion is animated by adjusting the opacity of one map over the other, and the two layers are pre-composed. This pre-composition can then be exported as an image sequence to drive the ray-tracing engine (see equation No. \vref{eq:ray map}).

On the pre-composed layer, the \textit{PixMap} effect can be applied to sample the \textit{footage map}. Following the previous steps, the warp-sampled footage can then be finalized to achieve the desired look.

\paragraph{Ray-Tracing with Distortion}\label{par:ray-tracing example}

In our 3D software, here \textit{Blender}, we import the animated \textit{view map} sequence and set-up nodes to convert the map coordinates into ray vectors (see equation No. \vref{eq:ray map}). These ray vectors are used to drive the \textit{Ray Portal BSDF} node.\cite{BlenderManual2024RayPortalBSDF}

The final 3D render can then be directly ray-traced with the embedded distortion, resulting in an animated lens that matches the warped footage from the compositing software.

\begin{note}
\label{note:4xMSAA}
To enhance the quality of footage warping, 4$\times$ MSAA filtering can be applied. This process involves pre-composing the animated \textit{view map} with the \textit{footage map} and increasing the resolution of the pre-composition by a factor of 4. The final \textit{PixMap} effect, used for sampling the footage, is then applied to this high-resolution pre-composition.
\end{note}

\subsection[Synthetic View Parameters]{Choosing Synthetic View Parameters}\label{subsec:sythetic parameters}

Understanding how distortion impacts the viewer's experience is crucial. Film directors typically select lenses based on their familiarity with specific lens characteristics and by conducting test shots using various options.\cite{Giardina2016AnamorphicUltra70,Sasaki2017Anamorphic,Neil2004CineLens}

By carefully choosing the synthetic projection, we can parameterize the viewing experience and make deliberate decisions about the visual feel and aesthetic of the image. Each type of lens projection imparts a distinct sensation to the viewer. Using the aximorphic model, it is possible to combine multiple projection characteristics into a single, non-animated distortion.

The table No. \vref{tab:perception} illustrates the typical sensory associations linked to the $k$ value from the aximorphic model.

\begin{table}[h]
	\centering
	\begin{tabular}{rll}
		\toprule
		\multicolumn{2}{c}{$k$ Value \& Projection Type} & Perception of Space \\\midrule
		$1$                & Rectilinear                       & straight lines \\
		$\nicefrac{1}{2}$  & Stereographic                     & shape, angles, space \\
		$0$                & Equidistant                       & speed, aim \\
		$-\nicefrac{1}{2}$ & Equisolid                         & distance, size \\
		$-1$               & Orthographic \textit{(fisheye)} & brightness \textit{(no vignetting)} \\
		\bottomrule
	\end{tabular}\\
	\footnotesize\emph{Source:} Empirical self study.
	\smallskip
	\caption{Projection type and corresponding perception of space, combined with $k$ value.}
	\label{tab:perception}
\end{table}

\paragraph{Exercise No. 1}\label{par:exercise car bumper camera}

Consider a dynamic shot with a camera mounted on the bumper of a car. The goal is to convey the speed of the car, the approaching distance to the target ahead, and the spatial awareness of the surroundings.

To achieve this, we can divide the frame into three regions: the bottom of the frame, showing the asphalt moving beneath the car; the top half of the frame, where the target appears to approach the car; and the sides of the frame, depicting the surrounding environment. Different $k$ values can be assigned to each region based on the information in Table \ref{tab:perception}, utilizing the aximorphic camera model.

\begin{itemize}[label=-]
	\item For the asphalt bottom region, we assign $k_\downarrow 0$ to enhance the sensation of speed.
	\item For the top half of the frame, where the target is approaching, we assign $k_\uparrow -\nicefrac{1}{2}$ to provide a sense of distance.
	\item For the sides of the frame, we assign $k_x \nicefrac{1}{2}$ to enhance the perception of space.
\end{itemize}

This approach allows for a nuanced manipulation of the viewer's sensory experience, tailored to the unique characteristics of each frame region.

\paragraph{Exercise No. 2}\label{par:exercise wide static shot}

Consider a generic scene featuring a family seated around a table. The goal is to capture the entire scene using a super wide-angle shot, while avoiding any distortion of the actors' faces. The camera is stationary on the table, and the family members are positioned around it. For this task, we select a projection that emulates the aesthetics of anamorphic photography.

\begin{itemize}[label=-]
	\item For the horizontal axis of the frame, we assign a $k_x$ value of $\nicefrac{1}{2}$ to preserve the proportions of the actors' faces.
	\item To minimize the distracting effects of the wide-angle shot and maintain focus on the performance, we can straighten the vertical lines of the shot by choosing a $k_y$ value of $0.88$, which closely resembles a rectilinear projection.
\end{itemize}

This configuration ensures the scene retains a natural and engaging appearance, emphasizing the performance while leveraging the expansive field of view provided by the wide-angle lens.

\begin{note}
	Both exercises can be accomplished using a wide-angle fisheye lens, with the desired look tailored during the post-production process by incorporating the LDES workflow.
\end{note}

\section[Future Work]{Future Work for Version 1.0}\label{sec:future}

Future work on this subject will include an evaluation of the calibration methods mentioned in Subsection \vref{subsec:calibration}. Further areas of exploration include:

\begin{itemize}
	\item \textbf{Real-Time Applications:} Explore the implementation of LDES in real-time systems, such as Virtual Production and in-camera special effects, where lens distortion correction and projection switching need to occur dynamically.
	\item \textbf{Integration with Rendering Pipelines:} Evaluate the application of LDES in rendering pipelines for visual effects, gaming, and 3D applications, including other ray-tracing workflows.
	\item \textbf{Subjective User Studies:} Conduct studies to evaluate how different synthetic distortions and transitions affect viewer perception, helping to refine the design of view maps for storytelling or aesthetic purposes.
	\item \textbf{Cross-Platform Compatibility:} Ensure compatibility of LDES across various software ecosystems, such as compositing, rendering, and editing tools.
	\item \textbf{Library of Lens\,\textit{View Maps}:} Develop an extensive, standardized library of lens \textit{view maps} for a variety of lens types, focal lengths, and distortion characteristics. This library could serve as a shared resource for professionals in filmmaking and visual effects, streamlining workflows and increasing creative flexibility.
\end{itemize}

These directions aim to broaden the utility and applicability of LDES while addressing potential limitations, and advancing the field of lens distortion modeling, directing, and correction.






\label{sec:reference}
\bibliographystyle{ACM-Reference-Format}
\bibliography{bibliography} 





\end{document}